\documentclass[10pt,twocolumn,letterpaper]{article}

\usepackage{cvpr}              

\usepackage{graphicx}
\usepackage{amsmath}
\usepackage{amssymb}
\usepackage{booktabs}

\usepackage{color}
\usepackage{xcolor}
\usepackage[dvipsnames]{xcolor}
\usepackage{colortbl}
\usepackage{bm}
\usepackage{listings}

\definecolor{orange_bar}{RGB}{244, 177, 131}
\definecolor{grey_bar}{RGB}{132, 151, 176}

\definecolor{darkgreen}{rgb}{0.0, 0.75, 0.0}
\definecolor{lightgreen}{rgb}{0.4, 0.4, 0.4}

\newcommand{\gdelta}[1]{{\footnotesize\textcolor{darkgreen}{#1}}}

\usepackage[pagebackref,breaklinks,colorlinks]{hyperref}

\usepackage[capitalize]{cleveref}
\crefname{section}{Sec.}{Secs.}
\Crefname{section}{Section}{Sections}
\Crefname{table}{Table}{Tables}
\crefname{table}{Tab.}{Tabs.}

\usepackage{fontawesome5}
\usepackage{hyperref}

\usepackage{graphicx}
\usepackage{caption}
\usepackage{float}      
\usepackage{xcolor}
\usepackage{minted}     
\usepackage{newtxtt}    

\setminted{%
  style=monokai,           
  bgcolor=black!2,  
  fontsize=\small,
  linenos=false,
  autogobble=true,
  breaklines=true,
  tabsize=2
}

\def\cvprPaperID{3016} 
\def\confName{CVPR}
\def\confYear{2023}

\begin{document}

\title{Dexbotic: Open-Source Vision-Language-Action Toolbox}

\def\huggingface{\raisebox{-1.5pt}{\includegraphics[height=1.05em]{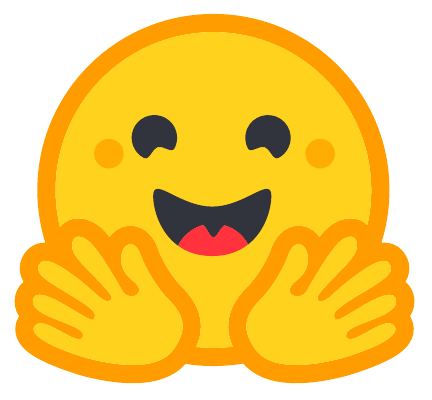}}}
\def\github{\raisebox{-1.5pt}{\includegraphics[height=1.05em]{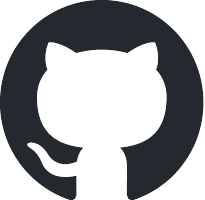}}}
\def\project{\raisebox{-1.5pt}{\includegraphics[height=1.2em]{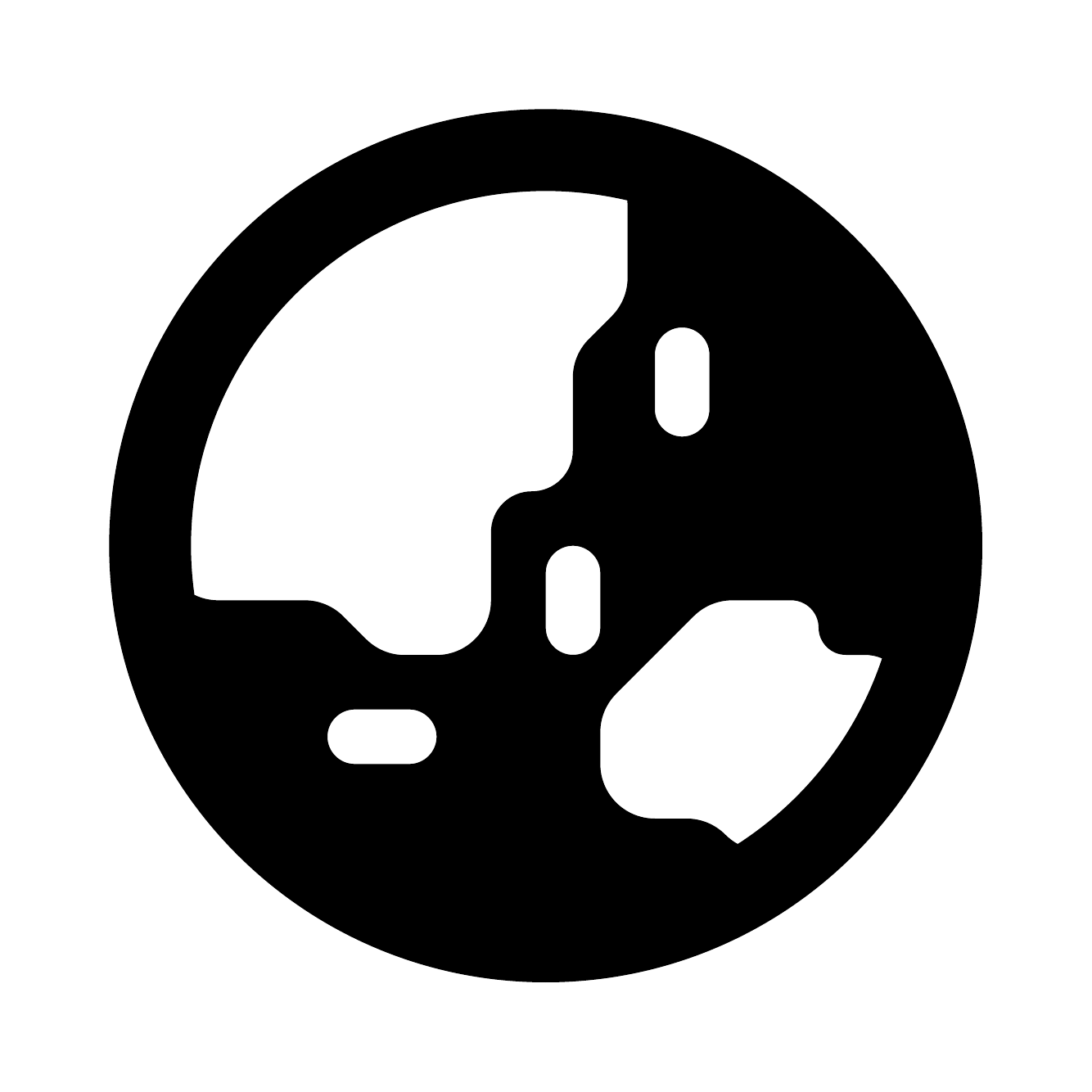}}}

\author{
  Bin Xie$^{1}$\thanks{Authors are listed in alphabetical order. $\dag$ Project lead.} \; Erjin Zhou$^{1}$ \; Fan Jia$^{1}$ \;  Hao Shi$^{1}$ \; Haoqiang Fan$^{1}$ \; Haowei Zhang$^{1}$ \; Hebei Li$^{1}$ \; \\ Jianjian Sun$^{2}$ \; Jie Bin$^{1}$ \; Junwen Huang$^{1}$ \;  Kai Liu$^{1}$ \;  Kaixin Liu$^{1}$ \; Kefan Gu$^{1}$ \;  Lin Sun$^{1}$ \; \\ Meng Zhang$^{1}$ \; Peilong Han$^{1}$ \; Ruitao Hao$^{1}$ \;  Ruitao Zhang$^{1}$ \; Saike Huang$^{1}$ \;  Songhan Xie$^{1}$ \; \\ Tiancai Wang$^{1\dag}$ \; Tianle Liu$^{1}$ \;  Wenbin Tang$^{1}$ \;  Wenqi Zhu$^{1}$ \; Yang Chen$^{1}$ \;  Yingfei Liu$^{1}$ \; \\ Yizhuang Zhou$^{2}$ \; Yu Liu$^{1}$ \; Yucheng Zhao$^{1}$ \;  Yunchao Ma$^{1}$ \;  Yunfei Wei$^{1}$ \;  Yuxiang Chen$^{1}$ \; \\ Ze Chen$^{1}$ \;  Zeming Li$^{2}$ \;  Zhao Wu$^{1}$ \;  Ziheng Zhang$^{1}$ \; Ziming Liu$^{1}$ \; Ziwei Yan$^{1}$ \; Ziyu Zhang \\
  $^1$ Dexmal \qquad $^2$ StepFun \\
  \small \project~\textbf{Project}: \url{https://dexbotic.com} \quad \\
  \small \github~\textbf{Github}: \url{https://github.com/Dexmal/dexbotic} \\
  \small \huggingface~\textbf{Huggingface}: \href{https://huggingface.co/collections/Dexmal/dexbotic-68f20493f6808a776bfc9fc4}{Dexbotic Collection} 
}

\maketitle

\begin{figure*}[t]
  \centering
  \includegraphics[width=\linewidth]{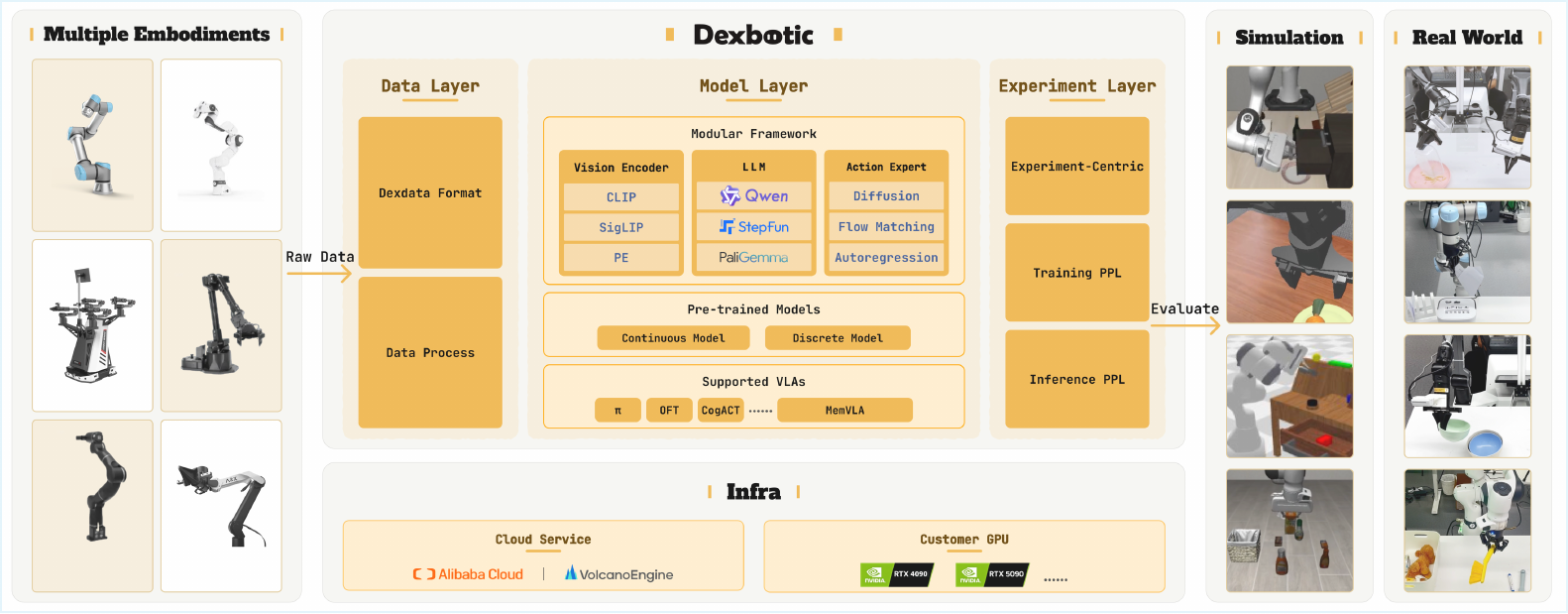}
  \caption{The overall architecture of Dexbotic. It introduces the Dexdata format to unify different embodiments. In Model Layer, Dexbotic integrates the open-source vision encoder, LLM and action expert through a unified modular VLA framework. Based on the provided DexboticVLMs, users can develop existing VLA policies and custom policies. Based on the developed policies, we further propose the Experiment Layer for fast development. Both the training pipeline and inference service are supported on some cloud service and customer GPUs.}
  \vspace{-2ex}
  \label{fig:teaser}
\end{figure*}

\begin{abstract}
In this paper, we present Dexbotic, an open-source Vision-Language-Action (VLA) model toolbox based on PyTorch. It aims to provide a one-stop VLA research service for professionals in the field of embodied intelligence. It offers a codebase that supports multiple mainstream VLA policies simultaneously, allowing users to reproduce various VLA methods with just a single environment setup. The toolbox is experiment-centric, where the users can quickly develop new VLA experiments by simply modifying the \textit{Exp} script. Moreover, we provide much stronger pretrained models to achieve great performance improvements for state-of-the-art VLA policies. Dexbotic will continuously update to include more of the latest pre-trained foundation models and cutting-edge VLA models in the industry.
\end{abstract}

\section{Introduction}
\label{sec:intro}

Recently, significant progress has been made in the field of embodied intelligence with the development of Vision-Language-Action (VLA) models~\cite{kim2024openvla, sun2025geovla, black2024pi_0, shi2025memoryvla, wen2025llada, brohan2023rt}. However, research in this area is fragmented across various institutions, each using different deep learning frameworks and model architectures. This diversity creates challenges for users when comparing policies, as they must configure multiple experimental environments and data formats, making the VLA development process cumbersome. Additionally, ensuring that each policy being compared is optimized to its fullest potential is difficult, leading to unfair comparisons. Furthermore, VLA models have evolved alongside Vision-Language Models (VLMs). However, many existing VLA models~\cite{kim2025fine, li2024cogact} are built on outdated VLMs~\cite{touvron2023llama}, making most users fail to benefit from the latest advanced VLMs.

To address these challenges, we release the so-called Dexbotic Toolbox for the Embodied AI community to push VLA research forward. Reviewing the toolbox development in AI 1.0 Era~\cite{chen2019mmdetection, detectron2}, the first step is to unify the model architecture. This process involves standardizing the structures and designs of models, which facilitated easier sharing, comparison, and improvement of algorithms across the research community. For example, mmdetection~\cite{chen2019mmdetection} disassembles the object detectors into backbone, neck and head. Similarly, existing VLA policies are uniformly divided into two parts in Dexbotic: vision-language model (VLM) and action expert (AE). The VLM part mainly includes a vision encoder, projector and large language model (LLM). It takes the observation and task prompt as inputs and produces the multi-modal tokens, which can be used to generate the discrete actions~\cite{kim2024openvla, brohan2023rt} or serves as the input of AE. The architecture of AE can be diffusion transformer~\cite{li2024cogact}, Multi-layer Perception (MLP)~\cite{kim2025fine} or Mixture-of-Experts (MoE)~\cite{black2024pi_0, black2024pi_05} coupled with LLM. 

Based on the unified model architecture above, Dexbotic further provides stronger pretrained models for some mainstream VLA policies, compared to the original open-source ones. Many existing VLA policies are built upon some outdated open-source VLMs or LLMs. For example, OpenVLA~\cite{kim2024openvla} and its follow-ups like CogACT~\cite{li2024cogact} and OFT~\cite{kim2025fine}, are all constructed based on Llama2~\cite{touvron2023llama}, whose representation is much inferior than those latest LLMs, like Qwen3~\cite{yang2025qwen3}. In Dexbotic, we introduce the DexboticVLMs, which integrates the open-source vision encoders with latest LLMs. Based on DexboticVLMs, we first provide discrete pretrained model for general VLM-based discrete policies. Users can utilize the discrete pretrained model for initialization of VLM part. For specific continuous-representation VLA policies, we further provide the detailed implementation for various action experts. Both the single-arm and hybrid-arm continuous pretrained models are provided to initialize the whole models.

To facilitate the development of VLA experiments, Dexbotic introduces an experiment-centric development framework. Different from those codebases that build upon the \textit{yaml} files for configuration~\cite{detectron2}, Dexbotic configurates the parameters by the \textit{Exp} scripts based on the provided \textit{base\_exp} script. User can simply modify these parameters that differentiates from the \textit{base\_exp} script without affecting the whole configuration. Users can easily meet various needs such as modifying configurations, changing models, or adding tasks by simply altering the \textit{Exp} script.
Moreover, Dexbotic supports the VLA training and inference on some cloud service like Alibaba Cloud as well as customer GPUs to satisfy the demand of different users. Dexbotic introduces the Dexdata format to support the training and deployment on multiple robots. The Dexdata format can save the storage for model training, compared to the LeRobot~\cite{cadene2024lerobot} and RLDS~\cite{ramos2021rlds} format.


\section{Main Features}
\label{gen_inst}

\subsection{Unified Modular VLA Framework}
Dexbotic centers around VLA models and is compatible with open-source interfaces of mainstream large language models (LLM). It integrates embodied manipulation and navigation, supporting multiple leading embodied manipulation and navigation policies, while also incorporating interfaces for future whole-body control.

\subsection{Powerful Pre-trained Foundation Models}
For mainstream VLA policies such as $\pi_0$~\cite{black2024pi_0} and CogACT~\cite{li2024cogact}, Dexbotic open-sources several more powerful pre-trained foundation models. These models bring significant performance improvements across various mainstream simulators, like SimplerEnv~\cite{li2024evaluating} and CALVIN~\cite{mees2022calvin}, as well as the real-world robotic tasks.

\subsection{Experiment-Centric Development Framework}
The experimental framework of Dexbotic adopts a "layered configuration + factory registration + entry dispatch" approach. Users can easily meet various needs such as modifying configurations, changing models, or adding tasks by simply altering the experimental Exp script. This design aligns with the Open-Closed Principle, allowing for flexibility and extensibility while maintaining stability.

\subsection{Cloud and Local Training Capabilities}
Dexbotic fully addresses the training needs of users from different universities and enterprises. It supports large-scale cloud-based training platforms such as Alibaba Cloud and Volcano Engine. Additionally, it accommodates local training with consumer-grade GPUs, such as RTX 4090 cards.

\subsection{Diverse Robot Training and Deployment}
For various mainstream robots, such as UR5, Franka and ALOHA, Dexbotic offers a unified data format for training. It also provides open-source, general-purpose deployment scripts, allowing users to customize their deployments. In the future, Dexbotic will continue to support additional mainstream robotic platforms.

\section{Supported VLA Policies}
\label{supported_vlas}
\noindent \textbf{Unified Policy Representation:}
\noindent The representation of different policies can be unified for both robotic manipulation and navigation, though their components may have some differences. Usually, the VLA policies can be simply divided into two parts: \textit{VLM} and \textit{Action Expert}.

\noindent \textbf{VLM} can be regarded as the backbone of VLA policy. It is usually pretrained on multimodality data, such as the image-text pairs for VQA and image caption. It mainly includes three parts: Vision encoder (e.g., CLIP~\cite{radford2021learning}, SigLIP~\cite{zhai2023sigmoid}) for visual token generation. Projector (e.g., two-layer MLPs) projects visual tokens into the textual space. LLM (e.g., Llama 2~\cite{touvron2023llama}) takes as input visual and textual tokens and produces tokens for text generation.

\noindent \textbf{Action Expert} takes the representation from VLM, such as the multi-modal tokens or cognition token, as input and produces the action chunking. For example, CogACT employs Diffusion Transformer while $\pi_0$ uses the flow matching. Currently, Dexbotic supports these VLA policies for robotic manipulation and navigation as follows. More VLA like \bm{$\pi_{0.5}$}~\cite{black2024pi_05} and VLN policies like NaVid~\cite{zhang2024navid} and NaVILA~\cite{cheng2024navila} will be supported in the near future.\\

\noindent $\bullet$ \bm{$\pi_0$}~\cite{black2024pi_0} is a well-known flow matching VLA policy and built upon the PaliGemma~\cite{beyer2024paligemma} and uses the action expert to generate action chunking. \\

\noindent $\bullet$ \textbf{OpenVLA-OFT}~\cite{kim2025fine} can be regarded as the improved version of OpenVLA~\cite{kim2024openvla} and explores fine-tuning strategies to greatly improve manipulation performance.\\

\noindent $\bullet$ \textbf{CogACT}~\cite{li2024cogact} is also based on the OpenVLA and extracts the cognition token, together which the noises are input to the Diffusion Transformer for diffusion modeling.\\

\noindent $\bullet$ \textbf{MemoryVLA}~\cite{shi2025memoryvla} further introduces the concept of perceptual-cognitive memory, improving the performance on long-horizon tasks. \\

\noindent $\bullet$ \textbf{MUVLA}~\cite{han2025muvla} is a recently proposed VLA policy based on map understanding for object navigation and achieves state-of-the-art performance.

\section{Architecture}
\label{arch} 

\subsection{Overall Architecture}
\begin{figure}[t]
  \centering
  \includegraphics[width=\linewidth]{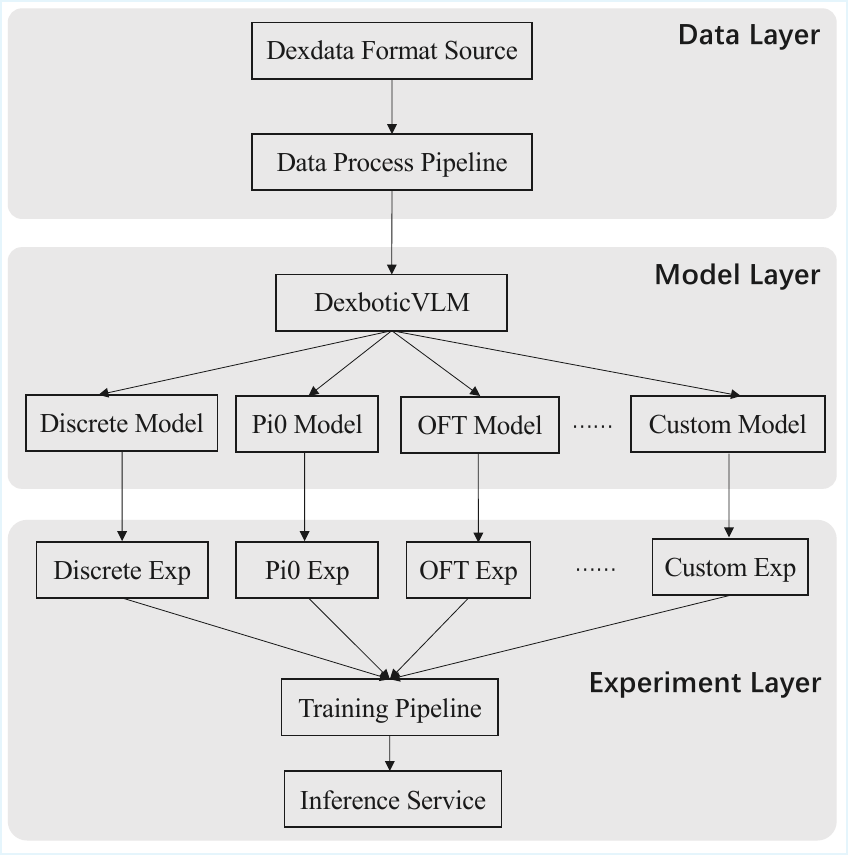}
  \caption{The overall architecture of Dexbotic. The framework is organized into three core layers including the data, model and experiment layers, that work together to provide a complete solution for training and serving VLA models.}
  \vspace{-2ex}
  \label{fig:overall_ppl}
\end{figure}

Fig.~\ref{fig:overall_ppl} shows the overall architecture of Dexbotic toolbox. It mainly includes three typical layers: Data Layer, Model Layer and Experiment Layer. In Data Layer (Sec.~\ref{data_layer}), we define the so-called Dexdata format to unify different data sources and save the storage. With Dexdata format data, dexbotic performs the data process to extract the image, text and state for training. In Model Layer (Sec.~\ref{model_layer}), we introduce the basic DexboticVLM as the foundation model to develop more VLA policies. DexboticVLM can be directly used for discrete VLA training, like RT-2~\cite{brohan2023rt} and OpenVLA~\cite{kim2024openvla}. It can also serve as the base model of existing VLA policies. For current version, we support multiple VLA policies like $\pi_0$, OpenVLA-OFT and CogACT. User can directly define their own custom VLA models. The most important part of dexbotic is the Experiment Layer (Sec.~\ref{exp_layer}). Based on different VLA models in Model Layer, we introduce various experiment scripts to support fast development for existing and custom policies while maintaining the stability of whole toolbox. With the experiment scripts, users can perform the training pipeline and the inference service using different modes.

\subsection{Data Layer}
\label{data_layer}
\begin{figure}[ht]
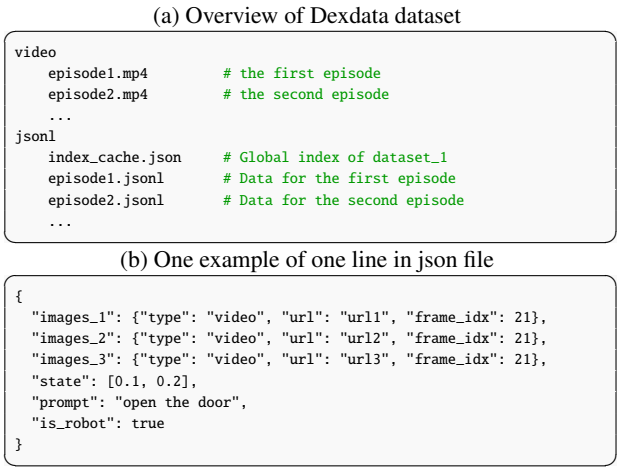

\centering
\vspace{-0.6em}

\begin{minipage}{0.45\textwidth}
\centering
{\small(a) Overview of Dexdata dataset}
\vspace{-0.4em}
\begin{lstlisting}[language=jsonl, linewidth=\linewidth]
video
    episode1.mp4         # the first episode
    episode2.mp4         # the second episode
    ...
jsonl
    index_cache.json     # Global index of dataset_1
    episode1.jsonl       # Data for the first episode 
    episode2.jsonl       # Data for the second episode 
    ...
\end{lstlisting}
\vspace{-0.4em}
\end{minipage}
\hfill
\begin{minipage}{0.45\textwidth}
\centering
{\small(b) One example of one line in json file}
\vspace{-0.4em}
\begin{lstlisting}[language=jsonl, linewidth=\linewidth]
{
  "images_1": {"type": "video", "url": "url1", "frame_idx": 21},
  "images_2": {"type": "video", "url": "url2", "frame_idx": 21},
  "images_3": {"type": "video", "url": "url3", "frame_idx": 21},
  "state": [0.1, 0.2],
  "prompt": "open the door",
  "is_robot": true
}
\end{lstlisting}
\end{minipage}

\vspace{-0.6em}
\caption{The overview of Dexdata format.}
\label{fig:dexdata_format}
\end{figure}

\noindent \textbf{Dexdata Format:}
We designed the Dexdata format to store robotic datasets in a unified and efficient way. As shown in Fig.~\ref{fig:dexdata_format} (a), it mainly includes two elements: \textit{video} and \textit{jsonl}. The \textit{video} directory contains video files with mp4 format while \textit{jsonl} directory includes the corresponding jsonl files. Each jsonl file contains the data for a single robot episode. In \textit{jsonl} directory, the \textit{index\_cache.json} file, which users can ignore, stores the metadata for all episodes and is automatically generated for fast access. Fig.~\ref{fig:dexdata_format} (b) illustrates one example of one line in a given jsonl file. It provides the detailed information of a frame about the multi-view images, robot state and the textual prompt. For the detail of data process pipeline, please kindly refer to the Sec.~\ref{exp_layer}.


\lstdefinelanguage{jsonl}{
  morekeywords={},
  sensitive=true,
  morecomment=[l]{\#},
  morestring=[b]"
}
\lstset{
    language=jsonl,
    basicstyle=\fontsize{6}{8}\ttfamily\selectfont,
    breaklines=true,
    frame=single,
    backgroundcolor=\color{lightgray!10},
    commentstyle=\color{green!60!black},
    keywordstyle=\color{blue},
    stringstyle=\color{black},
    frameround=tttt,
    columns=fullflexible,
    keepspaces=true,
    showstringspaces=false,
    xleftmargin=0pt,
    xrightmargin=0pt
}



\subsection{Model Layer}
\label{model_layer}
In this section, we first introduce our pretrained VLM, called DexboticVLM. Based on such a foundation model, we further describe how to develop existing and custom VLA policies. After that, we introduce some robotic pretrained models for finetuning both manipulation and navigation tasks.

\subsubsection{DexboticVLM}
For better generalization and the best use of existing open-sourced state-of-the-art LLM, we pretrain VLM from scratch.
In Dexbotic, we choose to pretrain our own VLMs, called DexboticVLM. 
We utilize the CLIP~\cite{radford2021learning} as the vision encoder, two-layer MLP as the projector and Qwen2.5 as the LLM. Similar to the LLaVA training pipeline~\cite{liu2023visual}, we first freeze the vision encoder and LLM parts and only train the projector for cross-modal alignment. After that, the parameters of whole network, including the vision encoder, projector and LLM, are updated. The training data sets include those of LLaVA and Cambrian~\cite{tong2024cambrian}.

\subsubsection{Model Development}
As described in Sec.~\ref{supported_vlas}, existing VLA pipeline can be unified with VLM and action expert parts. The DexboticVLM mentioned above can be directly used for discrete VLA training, the same as~\cite{kim2024openvla, brohan2023rt}. To enable DexboticVLM to directly predict robot actions, the actions in LLM output space are represented by mapping continuous robot actions to discrete tokens used by the language model tokenizer. Each dimension of the robot actions is discretized separately into 256 bins. To reproduce the existing continuous representation VLA models, we can build different action experts to produce various policies. For example, we can add the Diffusion Transformer as the action expert to construct the CogACT~\cite{li2024cogact} model. We can further introduce the memory module between DexboticVLM backbone and action expert to produce MemoryVLA~\cite{shi2025memoryvla} policy.
For current version, we support multiple VLA policies like $\pi_0$~\cite{black2024pi_0}, OpenVLA-OFT~\cite{kim2025fine} and CogACT. On the other hand, they can also develop their customized VLA policies by designing new action expert or introducing a supervision pipeline. 

\begin{figure}[t]
  \centering
  \includegraphics[width=\linewidth]{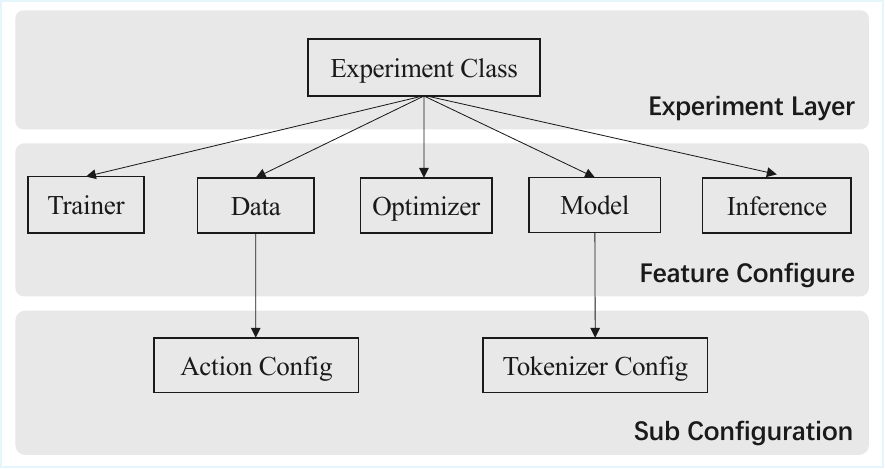}
  \caption{The layered configuration architecture in experiment layer. Each experiment class includes the configurations on trainer, data, optimizer, model and inference.}
  \vspace{-2ex}
  \label{fig:layered_config}
\end{figure}

\subsubsection{Pretrained Models}
For different demands on various robotic arm of users, we provide two kinds of pretrained models. The first one is the pretrained discrete model for general VLA policies and the second one is the pretrained continuous model for specific VLA policies. 
For continuous pretrained model, we further provide two versions for single-arm and two-arm tasks. We take CogACT as example to clearly illustrate the continuous pretrained models.

\noindent \textbf{Discrete Pretrained Model:}
Based on the DexboticVLM mentioned above, we further pretrain the discrete VLA model named \textit{Dexbotic-Base}. It is further pre-trained on single-arm data, which includes the subset of Open-X Embodiment dataset~\cite{o2024open}, simulation data from multiple simulators such as RLBench~\cite{james2020rlbench}, LIBERO~\cite{liu2023libero} and ManiSkill2~\cite{gu2023maniskill2}, and some real robot data (e.g. UR5). The DexboticVLM is trained to predict the $N$ discrete tokens to decode the discrete actions. Here, $N$ is the degree of freedom (DoF). During training, continuous actions of grounding truth are divided into 256 bins, which is used to supervise the predicted discrete tokens.
The pretrained discrete model \textit{Dexbotic-Base} can be directly used to finetune any VLM-based manipulation and navigation policies. The user can directly employ it for both discrete and continuous action learning. The parameters of VLM part can be loaded from the Dexbotic-Base checkpoint. For continuous action modeling, such as the diffusion~\cite{li2024cogact} and flow matching~\cite{black2024pi_0} processes, it can be performed by adding an action expert based on the DexboticVLM. The parameters of the action expert part can be randomly initialized.

\noindent \textbf{Single-arm Continuous Model:}
Here, we illustrate how to perform continuous pretraining for CogACT in Dexbotic.
Based on \textit{Dexbotic-Base}, we further train the entire CogACT model, including VLM and DiT head, through continuous representation pre-training. The VLM is initialized with \textit{Dexbotic-Base} and DiT head is randomly initialized. The GT continuous action is used to supervise the prediction of DiT. We use the data from multiple data resources, including the subset of Open-X embodiment dataset and part of our private dataset we collected. Our private dataset includes 52 manipulation tasks collected using eight single-arm real-world robots. These single-arm robots include UR5, Franka, Unitree Z1, Realman GEN72, Realman 75-6F, UMI, ARX5 and WidowX. Note that those robots have different embodiedments with different DoF, which challenges our infrastructure capacity. The pretrained model obtained is called \textit{Dexbotic-CogACT}.

\noindent \textbf{Hybrid-arm Continuous Model:}
The original CogACT policy~\cite{li2024cogact} does not support the multi-view and two-arm setting. To support two-arm tasks, we modify the number of noise tokens from 7 to 16, covering both 6-DoF and 7-DoF arms. The front-half tokens represent the actions of left arm while the second-half tokens denote the right arm one. We perform continue training on the robotic data of the hybrid arms based on the single-arm continuous model \textit{Dexbotic-CogACT} mentioned above. Despite the single-arm data above, we further introduce Robomind~\cite{wu2024robomind}, AgiBot World~\cite{bu2025agibot} datasets, as well as the private two-arm datasets collected from our ALOHA. For single-arm data, the single-arm actions are used to supervise the front-half tokens while the loss of second-half tokens is ignored during training. To support multi-view inputs, we share the vision encoder for multi-view images and their visual tokens are processed and concatenated as input for later LLM.

\subsection{Experiment Layer}
\label{exp_layer}
Experiment layer is the most important part in Dexbotic toolbox. In this section, we first describe how to develop new experiments from scratch. After that, the training pipeline and the inference service are presented in detail.

\subsubsection{Experiment Development}
Based on various models developed in Model Layer, we further develop the experiment scripts for model training and inference. As shown in Fig.~\ref{fig:layered_config}, we adopt a layered configuration architecture. For each experiment class, it contains the feature configurations on trainer, data, optimizer, model and inference. The data and model configurations further include the action and tokenizer configurations, respectively. We first construct a \textit{base\_exp} script in dexbotic, which serves as the basis configuration of VLA policies. This script mainly contains the configuration on optimizer, trainer, action, data, model and inference. 

To develop the experiment script for existing VLA policies, we inherit the \textit{base\_exp} script and modify the corresponding configuration for the target policy (e.g., \textit{CogACT\_Exp}). To run new experiments on existing policies, users can simply modify these configuration settings. As long as users inherit the corresponding configuration and override the fields, users can effortlessly fork a new set of hyperparameters without the need to copy the entire file. To develop the custom policies, users need to inherit and modify the configuration and model classes. 
Once the experiment script is correctly created, the users can directly run the experiment script to perform model training or inference like: \textit{python xxx\_exp.py \--task train}. Here, the \textit{task} denotes train or inference.

\begin{figure}[t]
  \centering
  \includegraphics[width=\linewidth]{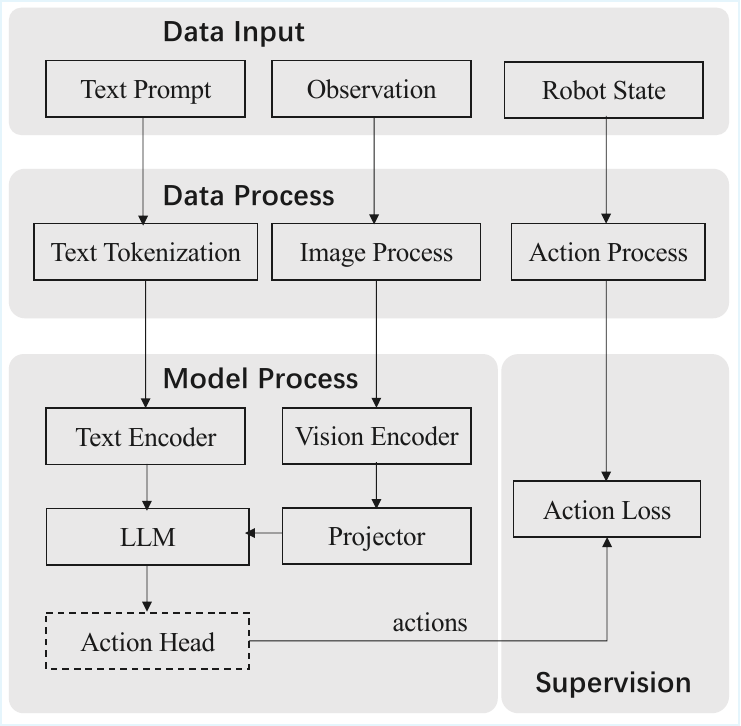}
  \caption{The training pipeline of Dexbotic.}
  \vspace{-2ex}
  \label{fig:training_ppl}
\end{figure}

\begin{figure}[t]
  \centering
  \includegraphics[width=\linewidth]{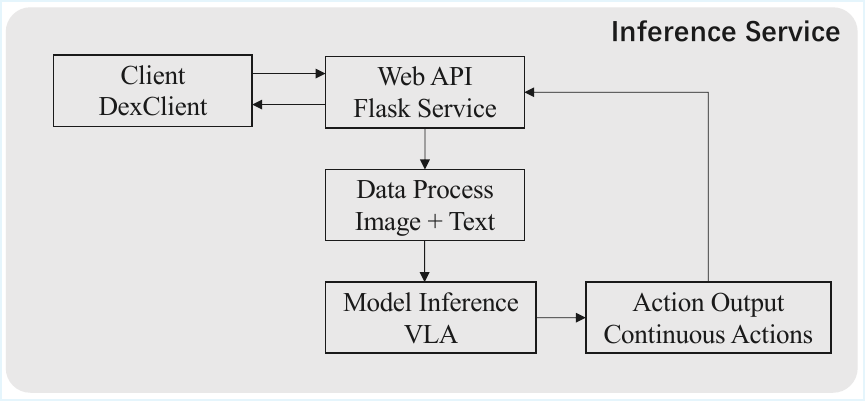}
  \caption{The inference service of Dexbotic.}
  \vspace{-2ex}
  \label{fig:inference_ppl}
\end{figure}

\begin{table*}[t]
  \centering
  \caption{Performance comparison on SimplerEnv-Bridge with WidowX robot between the state-of-the-art policies and dexbotic version.}
  \vspace{-1.6ex}
  \setlength{\tabcolsep}{1.6mm}{
  \begin{tabular}{l|ccccc}
    \toprule
    Methods       & Spoon on Towel & Carrot on Plate & Stack Cube & Eggplant in Basket & Avg. Suc \\\midrule
    CogACT        & 71.7  & 50.8  & 15.0  & 67.5  & 51.3    \\
    \rowcolor{gray!20} DB-CogACT     & 87.5  & 65.3  & 29.2 & 95.8 & \bf 69.5 \gdelta{(+18.2)} \\\midrule
    OFT           & 12.5  & 4.2  & 4.2 & 100.0 & 30.2 \\
    \rowcolor{gray!20} DB-OFT        & 91.7  & 76.4  & 43.1 & 94.4 & \bf 76.4 \gdelta{(+46.2)} \\\midrule
    MemVLA      & 75.0  & 75.0  & 37.5 & 100.0 & 71.9 \\
    \rowcolor{gray!20}DB-MemVLA   & 100.0  & 66.7  & 70.8 & 100.0 & \bf 84.4 \gdelta{(+12.5)}\\
    \bottomrule
  \end{tabular}
  }
  \label{tab:simplerenv}
\end{table*}

\begin{table*}[t]
  \centering
  \caption{Performance comparison on four tasks in RoboTwin2.0 between the state-of-the-art policies and their dexbotic version.}
  \vspace{-1.6ex}
  \setlength{\tabcolsep}{1.6mm}{
  \begin{tabular}{l|ccccc}
    \toprule
    Methods       & Adjust Bottle & Grab Roller & Place Empty Cup & Place Phone Stand & Avg. Suc\\\midrule
    CogACT        & 87  & 72  & 11  & 5  & 43.75    \\
    \rowcolor{gray!20}DB-CogACT     & 99  & 89  & 28 & 18 & \bf 58.5 \gdelta{(+14.75)} \\
    \bottomrule
  \end{tabular}
  }
  \label{tab:robotwin}
\end{table*}

\begin{table}[t]
  \centering
  \caption{Performance comparison on LIBERO between the state-of-the-art policies and their dexbotic version.}
  \vspace{-1.6ex}
  \setlength{\tabcolsep}{1.2mm}{
  \begin{tabular}{l|ccccc}
    \toprule
    Methods       & Spatial & Object & Goal & Long & Avg. Suc \\\midrule
    CogACT     & 97.2  & 98.0  & 90.2  & 88.8 & 93.6    \\
    \rowcolor{gray!20}DB-CogACT  & 93.8  & 97.8  & 96.2 & 91.8 & \bf 94.9 \gdelta{(+1.3)} \\ \midrule
    MemVLA      & 98.4  & 98.4  & 96.4 & 93.4 & 96.7 \\
    \rowcolor{gray!20} DB-MemVLA   & 97.2  & 99.2   & 98.4  & 93.2  & \bf 97.0 \gdelta{(+0.3)} \\
    \bottomrule
  \end{tabular}
  }
  \label{tab:libero}
\end{table}

\subsubsection{Training Pipeline}
As shown in Fig.~\ref{fig:training_ppl}, the data inputs of Dexbotic include the observation, textual instruction and robot states. The textual prompt is tokenized and input to the text encoder to generate the text tokens. The observation image is firstly processed by the vision encoder to generate the image tokens and aligned to the text space by a light-weight MLP-based projector. The image and text tokens are concatenated together and further input to the LLM to generate discrete tokens. For discrete-representation policy, these tokens can be directly decoded into sparse actions while for the continuous representation, an action expert is usually appended to produce continuous-value action chunking. The generated action sequence is supervised by the ground-truth of actions using the corresponding action losses.

\subsubsection{Inference Service}
Dexbotic also provides the inference service for different developers. As shown in Fig.~\ref{fig:inference_ppl}, DexClient sends a request to the Web API over the network. Web API based on the Flask Service, receives and process the request from DexClient. After receiving data from the DexClient, the system performs the data process on image and text, making the data suitable for next VLA model inference. The VLA model takes the image and text prompt as input and produces the continuous actions. The generated action sequences are then sent back to the Web API and Client sequentially. The DexClient performs corresponding actions based on these resulting actions.

\section{Benchmarks}
To validate the effectiveness of pretrained models we provided, we conduct the experiments and perform performance comparison on multiple simulation benchmarks: LIBERO~\cite{liu2023libero}, SimplerEnv~\cite{li2024evaluating}, CALVIN~\cite{mees2022calvin}, ManiSkill2~\cite{gu2023maniskill2}, Robotwin2.0~\cite{chen2025robotwin}. We first simply describe these simulation benchmarks and then show detailed benchmarking results on them. 

\begin{table*}[t]
  \centering
  \caption{Performance comparison on CALVIN between the state-of-the-art policies and their dexbotic version. We perform experiments on the ABC→D split, where VLA models are trained with data from A, B, and C environments and evaluated in environment D that is unseen during training.}
  \vspace{-1.6ex}
  \setlength{\tabcolsep}{2.4mm}{
  \begin{tabular}{l|cccccc}
    \toprule
    Methods       & 1 & 2 & 3 & 4 & 5 & Avg. Len \\ \midrule
    CogACT     & 0.838  & 0.729  & 0.640  & 0.559 & 0.480 & 3.25    \\
    \rowcolor{gray!20} DB-CogACT  & 0.935  & 0.867  & 0.803 & 0.760 & 0.698 & \bf 4.06\gdelta{(+0.81)} \\ \midrule
    OFT        & 0.891  & 0.794  & 0.674 & 0.598 & 0.515 & 3.47 \\
    \rowcolor{gray!20} DB-OFT     & 0.928  & 0.807 & 0.692  & 0.602 & 0.511 & \bf 3.54\gdelta{(+0.07)} \\ 
    \bottomrule
  \end{tabular}
  }
  \label{tab:calvin}
\end{table*}

\begin{table*}[t]
  \centering
  \caption{Performance comparison on five tasks in ManiSkill2 between the state-of-the-art policies and their dexbotic version.}
  \vspace{-1.6ex}
  \setlength{\tabcolsep}{1.2mm}{
  \begin{tabular}{l|cccccc}
    \toprule
    Methods       & PickCube & StackCube & PickSingleYCB & PickSingleEGAD & PickClutterYCB & Avg. Suc\\\midrule
    CogACT        & 55  & 70  & 30  & 25 & 20 & 40    \\
    \rowcolor{gray!20}DB-CogACT     & 90  & 65  & 65 & 40 & 30 & \bf 58 \gdelta{(+18.0)} \\
    \midrule
    OFT        & 40  & 45  & 5  & 5 & 0 & 21    \\
    \rowcolor{gray!20}DB-OFT    & 90  & 75  & 55 & 65 & 30 & \bf 63 \gdelta{(+42.0)} \\
    \bottomrule
  \end{tabular}
  }
  \label{tab:maniskill2}
\end{table*}

\subsection{Simulation Benchmarks}
\noindent\textbf{SimplerEnv} aims to narrow the gap
between simulation and the real world environment. It provides two embodiedments, Google robot and WidowX robot, and two task suites, Visual Matching and Variant Aggregations, for fair evaluation. In this paper, we focus mainly on the WidowX robot with only Visual Matching suite. It includes four tasks: \textit{Put Spoon on Towel}, \textit{Put Carrot on Plate}, \textit{Stack Cube} and \textit{Put Eggplant in Yellow Basket}.

\noindent\textbf{CALVIN} targets long-horizon language-conditioned robot manipulation tasks. We perform experiments under the standard ABC-D setting. The model is trained on environments A, B, and C, and evaluated on generalization capability on environment D. We report the average success rate over 1000 rollouts per task, along with the average number of tasks completed consecutively to accomplish five instructions. 

\noindent\textbf{ManiSkill2} mainly focuses on the basic pick-and-place. We evaluated the experimental results on five representative tasks: PickCube, StackCube, PickSingleYCB, PickSingleEGAD, and PickClutterYCB. Those tasks require the robot to grasp the specific object and place it in a 3D position indicated by a green marker, evaluating 3D perception and spatial reasoning. 

\noindent\textbf{RoboTwin2.0} is a newly-introduced simulation benchmark. It improves sim-to-real transformation and contains 50 dual-arm tasks and five robot embodiments. In this paper, we conduct the comparison based on four carefully selected tasks: adjust bottle, grab roller, place empty cup, and place phone stand. 

\noindent\textbf{LIBERO} includes five task suites, and each suite is designed to evaluate specific capabilities. \textit{LIBERO-Spatial} mainly focuses on different positions to place the objects.
\textit{LIBERO-Object} involves pick and place various objects into the box within a fixed scene layout. \textit{LIBERO-Goal} evaluates the ability to perform various operations in a fixed layout.
\textit{LIBERO-Long}, also called \textit{LIBERO-10}, which targets 10 long-horizon tasks that involve various scenes and operations. \textit{LIBERO-90} is an expanded version of LIBERO-10, presenting a more challenging benchmark.

\subsection{Benchmark Results}

\noindent\textbf{SimplerEnv:} As shown in Tab.~\ref{tab:simplerenv}, on the challenging SimplerEnv benchmarks, DB-CogACT with the pretrained model outperforms the official CogACT by absolute 18.2\%. DB-OFT achieves absolute 46.2\% improvements compared to the official OpenVLA-OFT. Moreover, we further evaluate the effectiveness of pretrained models on the MemoryVLA, a state-of-the-art VLA on SimplerEnv. The experimental result shows DB-MemoryVLA achieves 84.4\% success rate, achieving more than absolute 12\% improvements. The great improvements indicate the strong representation power of our pretrained models. 

\noindent\textbf{CALVIN:} To evaluate the improvements on long-horizon tasks, we conduct the performance comparison between VLA policies and their dexbotic countparts (see Tab.~\ref{tab:calvin}). DB-CogACT outperforms the official CogACT on all metrics. It achieves 4.06 average length, surpassing the CogACT by 0.81. Moreover, DB-OFT also achieves better performance than the standard OpenVLA-OFT. 

\noindent\textbf{RoboTwin 2.0:} Here we mainly take the CogACT as an example to show the effectiveness of dexbotic under the easy mode. As shown in Tab.~\ref{tab:robotwin}, for four selected tasks: adjust the bottle, grab roller, place empty cup and place the phone stand, CogACT achieves an average success rate of 43.75\%. In comparison, DB-CogACT surpasses CogACT by 14.75\% absolute gains with 58.5\% success rate. It demonstrates that our pretrained model can bring large performance improvements under the dual-arm embodiment. 

\noindent\textbf{LIBERO:} On LIBERO benchmark, the performance of state-of-the-art VLA policies is nearly saturated (see Tab.~\ref{tab:libero}). With our Dexbotic pre-trained models, those policies can obtain some further performance improvements for them like CogACT and MemoryVLA. Specifically, DB-CogACT boosts the average success rate on four task suites by 1.3\% points, compared to the CogACT baseline.

\noindent\textbf{ManiSkill2:} As shown in Tab.~\ref{tab:maniskill2}, the original OpenVLA-OFT produces undesirable performance with 21\% average success rate among five tasks. In comparison, DB-OFT improves the absolute performance by 42\% points, which demonstrates the effectiveness of our pretrained model. Moreover, DB-CogACT further improves the average success rate by 18\% points compared to the strong baseline of original CogACT.

\begin{figure*}[htbp]
  \centering
  \begin{minipage}{0.45\textwidth}
    \centering
    (a) Push the buttons
  \end{minipage}%
  \hfill
  \begin{minipage}{0.45\textwidth}
    \centering
    (b) Arrange fruits in basket
  \end{minipage}

  \includegraphics[width=\linewidth]{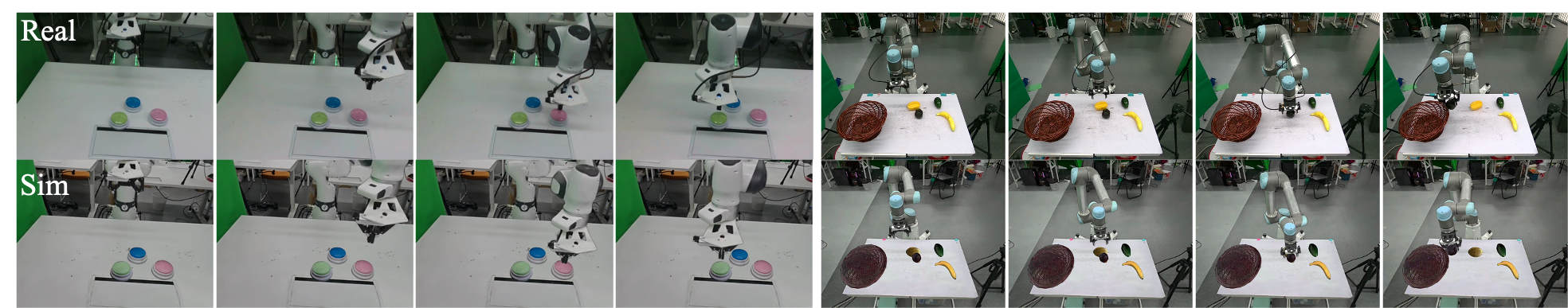}
  \caption{Comparison of real-world and simulated renderings, highlighting the high consistency in the robotic arm and objects performing the same action in both environments.}
  \vspace{-2ex}
  \label{fig:real2sim}
\end{figure*}

\begin{figure*}[t]
  \centering
  \includegraphics[width=\linewidth]{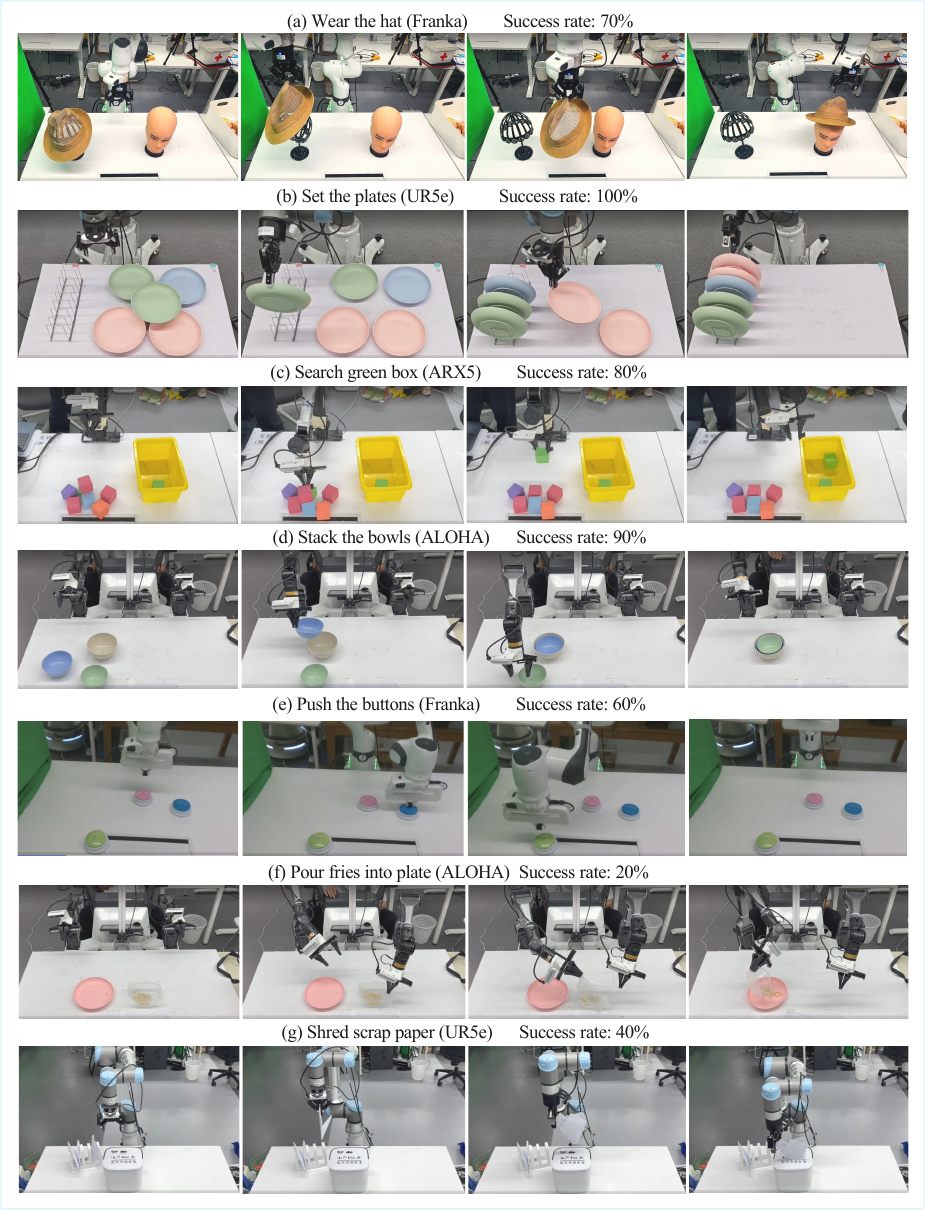}
  \caption{The video gallery produced by Dexbotic toolbox.}
  \vspace{-2ex}
  \label{fig:gallery}
\end{figure*}

\section{Real-world Performance}
To showcase what tasks users can accomplish in real world using the Dexbotic toolbox, we release the task gallery for visualization (see Fig.~\ref{fig:gallery}). On different robots like UR5e, ALOHA, ARX5 and Franka, we collect the real-world task data through the teleoperation. For each task, we collect 500-1000 demonstrations depending on the task difficulty and convert those data into our Dexdata format. These converted data is used to finetune the corresponding models based on our pretrained model. 

The real world experiments show that Dexbotic can accomplish various daily tasks. Notably, it achieves 100\% and 80\% success rates for the \textit{set the plates} and \textit{search the green box} tasks, respectively. However, for those fine-grained manipulation tasks like \textit{Shred the scrap paper} and \textit{Pour fries into plate}, they indeed pose challenges for existing VLA policies. Moreover, we verify that some state-of-the-art VLA policy like MemoryVLA~\cite{shi2025memoryvla} can solve the long-horizon and memory-requiring tasks, like \textit{Push buttons sequentially}.
Please see the \href{https://dexbotic.com}{official website} of Dexbotic for more visualization on real-world tasks, and we would like to encourage the users to utilize the Dexbotic toolbox to develop more real-world robotic tasks. We would also suggest users submit more policies developed based on Dexbotic to \href{https://robochallenge.ai/}{RoboChallenge} for fair comparison in real-world.

\section{Real2Sim Evaluation}

Real-world evaluations are often labor-intensive. To address this challenge, we propose Dexbotic Open Source-Twins (DOS-Twins), a Real2Sim2Real ~\cite{li2024robogsim,qureshi2025splatsim} simulator developed as part of the Dexbotic ecosystem. For the publicly released real-world datasets, we reconstruct corresponding simulation environments that closely replicate the real setups. Users can leverage these datasets for model training and subsequently submit their trained models to our Real2Sim evaluation interface for comprehensive capability assessment. DOS-Twins ensures consistency between simulation and real-world across three key dimensions:

\noindent\textbf{Visual Consistency.} The VLA model is highly sensitive to visual alignment. This misalignment leads to discrepancies in success rates, with high success in simulation but significantly lower success in real-world tasks. Leveraging 3D Gaussian Splatting (3DGS)~\cite{kerbl20233d}, we generate photorealistic renderings with precise alignment between rendered objects and their corresponding meshes. Accurate camera calibration guarantees that the simulated camera viewpoints perfectly match those of real-world cameras. 

\noindent\textbf{Motion Consistency.} Motion consistency is crucial for ensuring that the simulated and real-world control systems are aligned. Proper alignment prevents incorrect visual feedback and ensures that the model can reliably execute the intended actions, as predicted by the simulation, in real-world tasks. Without consistency in motion, the robotic arm may not perform as expected, leading to discrepancies between simulated success and real-world failure. The low-level controller of the robotic arm is calibrated to match the motion dynamics and kinematic characteristics of the real hardware.

\noindent\textbf{Interaction Consistency.} In grasping tasks, ensuring accurate interaction between the gripper and the object is critical, as the geometric and physical alignment directly impacts the model’s ability to perform tasks successfully. To achieve this, we perform high-precision 3D scanning of both the gripper and objects, ensuring that the geometric structure errors remain within a millimeter for the objects. This high level of precision minimizes interaction errors, maintaining millimeter-level accuracy during grasping. Additionally, we align the parallel structure of the gripper with that of the real hardware, ensuring consistent physical interactions between the gripper and objects. This consistency enables reliable performance in both simulated and real-world environments.

Specifically, DOS-Twins employs Isaac Sim as the backend physics engine and 3DGS as the rendering frontend, achieving high-precision reconstruction and visualization of robotic arms and their components. It supports multiple robotic arm and gripper configurations and allows rendering from any viewpoint depending on task requirements. All simulation assets are organized modularly, enabling users to reuse and customize them for specific tasks.
Moreover, users can follow our simulation construction workflow and utilize the provided developer tools to build new environments. These environments can automatically monitor model performance across various robotic platforms and task settings during pre-training.

As illustrated in Fig.~\ref{fig:real2sim}, we compare the simulated environment with the real setup through replay analysis. The results show that the manipulation behavior in our simulations closely aligns with real-world phenomena, demonstrating that users can train policies in the real world while conducting consistent evaluations within our simulation framework. Please access the official website for more visualization comparison on the visual, motion and interaction consistency.

{\small
    \bibliographystyle{ieee_fullname}
    \bibliography{egbib}
  }

\end{document}